\documentclass[conference]{IEEEtran}

\IEEEoverridecommandlockouts
\usepackage{times}

\usepackage{balance}
\usepackage{graphicx}
\usepackage{amsmath} 		
\usepackage{amssymb}  	
\usepackage{dsfont}
\usepackage{mathrsfs}
\usepackage{algorithmic}
\usepackage{algorithm}
\usepackage[usenames,dvipsnames]{color}
\usepackage[latin1]{inputenc}
\usepackage{subfig}
\usepackage{forloop}
\usepackage{ifthen}
\usepackage{intcalc}
\usepackage{tikz}

\newcommand {\mibf}[1] {\boldsymbol{#1}}

\newcommand {\quitado}[1] {}

\newcommand {\bmilista}                   
{\begin{list} {$\bullet$} {
 \setlength{\rightmargin}{0em}
 \setlength{\leftmargin}{2em}
 \setlength{\topsep}{1ex}
 \setlength{\parsep}{0ex}
 \setlength{\parskip}{0ex}
 \setlength{\itemsep}{0ex} 
 \setlength{\listparindent}{0em}
 \setlength{\labelsep}{0.5em}
 \setlength{\labelwidth}{1em}
}}
\newcommand {\emilista} {\end{list}}
\newsavebox{\ieeealgbox}

\begin{document}
\title{Ontological Physics-based Motion Planning for Manipulation}

\author{Muhayyuddin, Aliakbar Akbari and Jan~Rosell\\
Institute of Industrial and Control Engineering (IOC)\\
Universitat Politcnica de Catalunya (UPC) -- Barcelona Tech\\
Barcelona, Spain, jan.rosell@upc.edu \thanks{This work was partially supported by the Spanish Government through the projects \mbox{DPI2011-22471}, \mbox{DPI2013-40882-P} and \mbox{DPI2014-57757-R}. Muhayyuddin is supported by the Generalitat de Catalunya through the grant FI-DGR 2014.}}

\maketitle

\begin{abstract}

Robotic manipulation involves actions where contacts occur between the robot and the objects. In this scope, the availability of physics-based engines allows motion
planners to comprise dynamics between rigid bodies, which is necessary for planning this type of actions. However, physics-based motion planning is computationally
intensive due to the high dimensionality of the state space and the need to work with a low integration step to find accurate solutions.
On the other hand, manipulation actions change the environment and conditions further actions and motions. To cope with this issue, the representation of
manipulation actions using ontologies enables a semantic-based inference processe that alleviates the computational cost of motion planning.
This paper proposes a manipulation planning framework where physics-based motion planning is enhanced with ontological knowledge representation and reasoning.
The proposal has been implemented and is illustrated and validated with a simple example. Its use in grasping tasks in cluttered environments is currently under
development.
\end{abstract}

\section{Introduction}\label{s-Introduction}
Motion planning problems deal with computing the collision-free trajectory from a start to a goal state in the configuration space, that is the set of all
possible configurations of the robot \cite{Perez1983}. During the last decade, the field of motion planning has evolved from the basic geometric problems (such as the
piano mover's problem) to the kinodynamic motion planning problems for complex robotic systems \cite{sucanK2012}. The kinodynamic motion planning algorithms consider
both the kinematic (geometric) and the dynamic (differential) constraints while planning. The most significant class of motion planning methods
are those based on the sampling of the configuration space,  such as the \textit{Rapidly-exploring Random Trees} (RRTs \cite{lavalle2000}) and \textit{Kinodynamic Planning by Interior-Exterior Cell Exploration} (KPIECE \cite{csucan2010}). Both of these methods are sampling-based kinodynamic
motion planners for systems with differential constraints, and KPIECE is particularly suited for systems with complex dynamics.

Typically, all these kinodynamic algorithms focus on computing the collision-free trajectories in the configuration space. A new class of motion planning strategies,
known as physics-based, has recently emerged that focus not only on computing collision-free trajectories, but also on considering the purposeful
manipulation of  objects. Physics-based planning is an extension to the kinodynamic motion planing that brings up the interaction between rigid bodies in terms of
considering their dynamics \cite{donald1993}. Unlike what it is done in collision-free motion planning, physics-based motion planning involves the complex
interaction between rigid bodies and manipulation actions to achieve the desired goal \cite{zickler2010}, i.e. based on Newtonian physics the behavior of the bodies
resulting from physical interactions is simulated and taken into account in the planning decisions. These algorithms implicitly use the sampling based algorithms for
sampling the states. To evaluate the interaction between rigid bodies, the state propagator uses dynamic simulators like the Open Dynamic Engine (ODE \cite{OpenDE2007}).

A physics-based motion planner has been proposed in \cite{zickler2009} that uses non-deterministic tactics and skills modeled using a finite state machine (i.e. the rigid inter-body 
dynamical simulations are performed using a physics-based state and transition model). Their proposal is called Behavioral Kinodynamic Rapidly-exploring Random Trees and the Behavioral Kinodynamic Balance Growth Trees.
A different approach proposed a physics-based temporal projection to provide, for manipulation tasks, a reasoning capability on motion planning parameters \cite{mosenlechner2013} (the
study applied Bullet physics engine to reason about stability, visibility, and reachability).
Focused in grasping, a physics-based grasp planner has been proposed that computes the motions of the objects using a physics-based pushing analysis
\cite{dogar2012}. It simultaneously contacts and moves the obstacle in a controlled way for clearing the required path (to make it computationally
efficient the interactions of the robot with the objects are precomputed). In a similar line, a physics-based trajectory optimization approach is proposed in
\cite{kitaev2015} that samples the straight line trajectories and evaluates the cost based on various parameters such as collision costs, clutter behavior costs, and
trajectory matching costs.

None of these referred works, however, takes advantage of using reasoning on knowledge within the physics-based motion planning procedures, although knowledge-based
representation techniques have usually been used as one of the significant keys to model manipulation problems, i.e.
the integration of knowledge-based reasoning process with physics engines still is an open research challenge for manipulation problems, and this paper tries to
contribute in this line.
Regarding knowledge representation and reasoning, the use of ontologies has emerged as a good way to describe manipulation actions and to facilitate the reasoning process in manipulation planning to provide inference capabilities from the abstract knowledge \cite{feyzabadi2014, TaskPlanning2015}. With this in mind, the main purpose of this paper is to develop a
framework where physics-based motion planning is enhanced with ontological knowledge and reasoning (about manipulation actions as well as about the geometry of the
objects), i.e. the inferred knowledge is to be used to tune the physics-based motion planning. The proposed ontological physics-based motion planner is called  the
\textit{smart motion planner}.

After this introduction the paper is structured as follows. Section \ref{s-ProblemStatement} describes the problem statement and the solution overview,
Sections \ref{s-OntologyReps} and \ref{s-ReasoningProcess} present, respectively, the knowledge representation and the reasoning, Section \ref{s-Implementation}
presents the framework, the planning algorithm, and the simulation results, and finally Sections \ref{s-Conclusions} concludes the work.

\section{Problem Statement and Solution Overview} \label{s-ProblemStatement} 

\subsection{Problem statement}

Consider a motion planning problem where a robot must move in an environment with obstacles that can be fixed or manipulatable (i.e. obstacles that can be pushed away). Let $Q$ be the state space of the robot, $q_{init} \in Q$ the initial state and $Q_{goal}\in Q$ the goal region. The robot must move from
$q_{init}$ to $q_{goal}\in Q_{goal}$ avoiding collisions with fixed obstacles and, if necessary, pushing away manipulatable obstacles to clear the way to find a
solution path (for instance a manipulatable obstacle may be placed at the goal region thus preventing a collision-free path to exist).

\begin{figure}\label{f-1}
\begin{center}
\includegraphics[width=6.5cm]{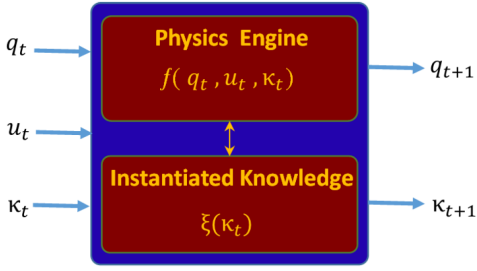}
\end{center}
\vspace{-3mm}
\caption{State transition process for ontological physics-based planning.}
\label{fig:Fig1}
\end{figure}

\subsection{Modelling}

We consider a workspace composed of rigid bodies, that are divided into two main categories, \textit{fixed bodies} and \textit{manipulatable bodies}. The former
will remain static during the whole planning process, even if  collisions occurs with other manipulatable objects. On the contrary,
the latter can be manipulated during the planning. The manipulatable bodies are further divided into \textit{free-manipulatable bodies} and \textit{constraint-oriented
manipulatable bodies}. Following the rigid body dynamics, the free-manipulatable bodies can move in any direction when collisions occur. On the contrary, constraint-oriented manipulatable bodies, have the allowable motion directions restrained, i.e. they can only be moved if forces are applied in certain
directions.

The manipulation constraints are modeled by defining some parts of the object from where the object can be pushed, and an associated region for each one (called manipulatable regions) where the robot must be located to exert the pushing forces. For instance, the manipulation constraints for a car-like body are defined by the forward and the backward
directions as the only allowable motions directions (i.e. the body will not move if it receives a contact force on its lateral side). These constraints are modelled by two parts (the rear and the front) and their manipulatable regions.


At any time $t$ the state $s$ of each body is defined as:
\begin{equation}
\label{state of a body}
	 s=\lbrace p, o, v, w, \eta \rbrace ,
\end{equation}
where $p$ represents the position of body, $o$  the orientation, $v$  the linear velocity, $w$  the angular velocity, and $\eta$ describes the manipulation
constraints.

Also, in order to model the problem, the knowledge about the task and the workspace is organized in two levels: a \textit{high level} knowledge representation called abstract knowledge $K$, and a \textit{low level} knowledge representation called \textit{instantiated knowledge} $\kappa_t$.
The former codes the semantic-based knowledge about the world, such as the type of bodies (i.e. fixed and manipulatable) and other
features like mass or the manipulation constraints in case of constraint-oriented manipulatable bodies. This knowledge remains unchanged
 throughout the whole planning process.
The latter, on the other hand, is a dynamic knowledge, i.e. the instantiated knowledge infers from the abstract knowledge and is continually updated by making use of the reasoning process (detailed in Section~\ref{s-ReasoningProcess}).
For instance, if at a given instance of time the car-like body has its front manipulatable region overlapping another manipulatable object, then this region cannot be accessed by the robot and the body cannot be pushed backward. In this case the front manipulatable region is tagged as not valid in the instantiated knowledge.

Then, the state of the world at any instant of time can be described as the tuple $(\kappa_t, \mathbf{q}_t)$, with $\kappa_t$ being the instantiated knowledge about the world at time $t$ and $\mathbf{q}_t$ the corresponding states of the bodies:
\begin{eqnarray}
\begin{split}\label{world state}
	 \mathbf{q}_t = \lbrace s_0,s_1, ..., s_n, t \rbrace,
\end{split}
\end{eqnarray}
This state evolves with the state transition process depicted in Fig. \ref{fig:Fig1}, that is
composed of two modules, the physics engine module and the instantiated knowledge module.
Given the current state $(\kappa_t, \mathbf{q}_t)$ and the controls $\mathbf{u}_t$ to be applied, the physics engine module implements the dynamic state propagator that computes $\mathbf{q}_{t+1}$ and the instantiated knowledge module is then used to update the current instantiated knowledge:
\begin{eqnarray}\label{eq:state propagator}
	\mathbf{q}_{t+1} &=& f(\mathbf{q}_t,\mathbf{u}_t,\kappa_t)\\
	 \kappa_{t+1} &=& \xi(\kappa_t)
\end{eqnarray}
This state transition process will be at the core of the physics-based planners to be used, and will only return the state whenever it is valid, i.e. the application of a control $\mathbf{u}_t$ that produces a non-allowable collision will not be accepted (collisions are not allowed with fixed obstacles nor with constraint-oriented manipulatable obstacles if the robot is not located at one of the manipulatable regions).

\subsection{Solution overview}

The problem will be solved with a physics-based motion planner enhanced with a knowledge-based reasoning process that on the one hand modifies how manipulatable objects behaves, and on the other modifies some parameters of the planner.
The key points regarding $\kappa_t$ and the reasoning process are:
\begin{itemize}
\item All fixed bodies are set as non-collisionable (the robot is not allowed to collide with them).
\item All free manipulatable bodies are set as collisionable.
\item Any constraint-oriented manipulatable body is set as collisionable if the robot is located at one of its manipulatable regions, otherwise it is set as non-collisionable.
\item If the manipulatable region of a constraint-oriented manipulatable body is occupied, then this region is set as not valid.
\item If a constraint-oriented manipulatable bodies result with no valid manipulatable regions, then the body is set as fixed.
\item If the state transition process results with a not-allowed collision, then the new generated state is discarded by the planner.
\item If the goal region is occupied by a constraint-oriented manipulatable body and the planner used has a sampling bias, then this sampling-bias is randomly set to one of the manipulatable regions of the body occupying the goal region, with the aim of pushing it away.
\end{itemize}

\section{Ontology-based manipulation representation} \label{s-OntologyReps}

Ontology models have recently been used to represent robotic manipulation specifications, like manipulation actions or object descriptions and their properties. These
models give robots the required knowledge to reason and to take better decisions while planning.
\begin{figure}[t]
\includegraphics[scale=0.55]{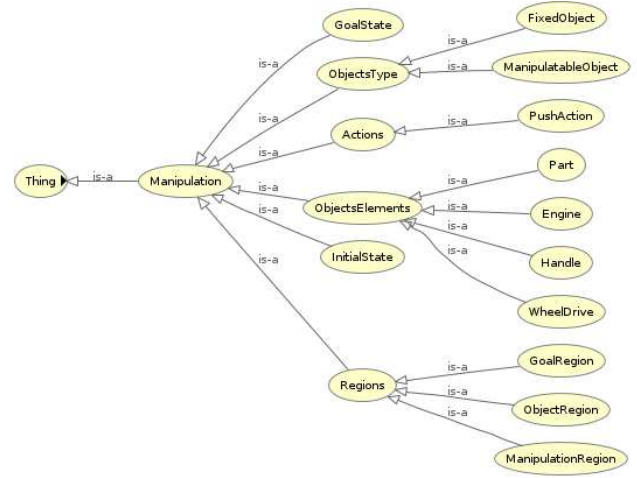}
\centering
\caption{Ontology-based manipulation classes.}
\label{fig:Fig2}
\end{figure}
\subsection{Ontology overview}
Ontology models organize knowledge within specific domains. They enable a flexible access by defining
things as well as their relations. Ontologies are composed of
 five components: 
 \begin{itemize}
\item \textit{Classes}, also referred to as a concepts, describe types or collections of objects that share common properties.
\item \textit{Individuals}, also called  instances, represent specific elements of classes, e.g., people are individuals of a person class.
\item \textit{Properties} express how classes and individuals are related to one another, e.g., sibling relationship among people.
\item \textit{Attributes} specify features, unique properties, and particular characteristics of objects, such as the age of a person.
\item \textit{Axioms} define constraints on the values of classes and individuals.
\end{itemize}
Ontologies can be encoded using the Web Ontology Language (OWL) \cite{owl2004}. OWL is intended to collect and organize ontology-based knowledge on a world-wide
accessible database in an XML-based file format in order that multiple systems can use and share such knowledge.

Ontologies can be written using the \textit{Prot\'eg\'e} editor  \cite{protege}. \textit{Prot\'eg\'e} is an open
source platform providing an ontology editor to develop knowledge-based applications. It, moreover, can be used to portray the visualization of ontology in terms of
showing relations of classes or individuals as a graph.

\subsection{Manipulation knowledge representation}
For representing a manipulation problem, it is required to specify actions, objects with their properties, initial and goal states of a robot, and valid regions. In
the present study, the \textit{Prot\'eg\'e} editor is employed to represent this knowledge in the OWL. Six classes, derived from a general \textit{``Manipulation''} class
 have been defined (Fig.~\ref{fig:Fig2}):
\begin{itemize} \itemsep1pt \parskip0pt \parsep0pt
  \item Class \textit{``InitialState''} contains the initial location of the robot.
  \item Class \textit{``GoalState''} contains the goal location of the robot.
  \item Class \textit{``Actions''} contains different manipulation actions, PickAction, PlaceAction, and PushAction, although only the latter one will be used in the
  present paper.
  \item Class \textit{``Regions''} defines different types of regions: ManipulationRegion, ObjectRegion, and GoalRegion. The ManipulationRegions are the regions
  around the constrained-oriented manipulatable bodies from where the interaction between the robot and the bodies are allowed (these regions are currently defined as   boxes), i.e. the robot can only interact with the bodies through the ManipulationRegions. ObjectRegion is the bounding box of an object. The GoalRegion is a circular region defined around the goal state.
  \item Class \textit{``ObjectsType''} collects the types of objects in the world: ManipulatableObject or FixedObject. If necessary, the manipulatable status of an
  object can be temporary changed to fixed by using the assertion process.
  \item Class \textit{``ObjectsElements''} describes elements and features of objects.
\end{itemize}

\section{Knowledge-based reasoning process} \label{s-ReasoningProcess}
Robots can be made more autonomous to carry out their complex tasks if some kind of reasoning process upon knowledge is provided, instead of having a purely symbolic
knowledge representation. Reasoning on manipulation actions or on robot geometries provides the possibility of understanding the tasks in an analogous way a human
does, e.g. of being able to answer questions like \textit{``Which specific actions should be selected for a given type of object?"}, or
\textit{``Do I need to move some manipulatable objects to reach the goal?"}.
\subsection{Types of reasoning process}
Two types of reasoning are proposed: reasoning about the manipulation actions and about the geometry of the objects.
The aim of the reasoning about the manipulation actions   is
to guide the motion planner to find an appropriate action in accordance with the object type.
As introduced before, objects are classified into \textit{fixed} and \textit{manipulatable}, and those manipulatable are divided into \textit{constraint-oriented
manipulatable} and \textit{free manipulatable}.  The type of object can be predefined, although it can be identified by a reasoning process based on the parts and
features of the object and on the constraints they define, since these constraints may restrict the movement of the object.
For instance, if when defining a car-like object the body is attached with a wheel-drive, the allowable directions of motion are automatically constrained along
the plane of the wheel  and  the pushing manipulation actions can then only be exerted from some parts of the object (the front and the rear).
Also, some parameters like the size and the weight of objects may constrain which actions can be selected according to the capabilities of a robot.

The reasoning about the geometry of objects tackles two matters. First, it determines whether the goal region is free or occupied by other objects (using the
bounding boxes of the objects and of the goal region). Second, it determines possible manipulation regions on the object (from where to
apply pushing forces). To address this, a reasoning process is done by inferring over the properties of the objects.
\begin{figure}
\begin{center}
\includegraphics[width=7.0cm]{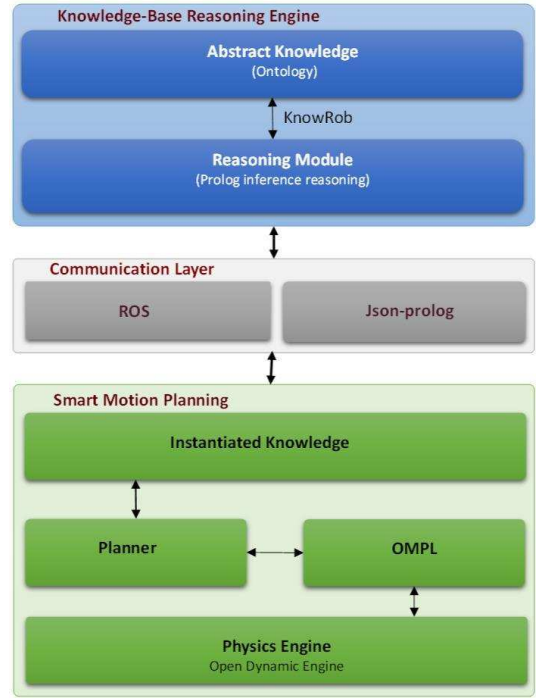}
\caption{Integration framework for ontological physics-based motion planning}\label{fig:Fig3}
\end{center}
\end{figure}

\subsection{Prolog-based knowledge inference process}
The reasoning predicates have been defined in Prolog and  been integrated within Knowrob,  a knowledge processing framework for robotic systems \cite{knowrob2009}.
Knowrob is a powerful reasoning tool that works over ontology models: its implementation is based on SWI Prolog and the Semantic Web library, enabling the use of
Prolog predicates to fetch the knowledge stored inside the OWL. Therefore, Knowrob  provides a flexible access to such knowledge
to facilitate the reasoning process.

The following Prolog predicates have been implemented  to infer manipulation knowledge from the OWL and to reason according to the states of the bodies. 
\begin{itemize}
\item
\textit{object\_classification(?Obj, ?ObjType, ?ManipType)}: Given an object \textit{Obj} returns its type (fixed or manipulatable) in \textit{ObjType}.
In case of a manipulatable objects, its type (free or constrained) is inferred from its properties and returned in \textit{ManipType}.
\item \textit{action\_type(?Obj, ?ObjElement, ?Action)}: Given an object \textit{Obj} returns in \textit{ObjElement} a list of elements (parts) from where the object can be manipulated using the list of actions returned in \textit{Action}, e.g. for the car-like object it will return the rear and the front as elements and push as the manipulation action for both.
\item \textit{determine\_goal\_region(?Obj)}: Returns in \textit{Obj} the object that is occupying the goal region if any, or null otherwise.
\item \textit{valid\_region(?Obj, ?ObjElement, ?ValidRegion)}: Given an object \textit{Obj} and its part \textit{ObjElement} returns in  \textit{ValidRegion} whether the associated  manipulatable region is valid or not.
\end{itemize}

\section{Implementation} \label{s-Implementation}
\subsection{Framework} \label{ss-Framework}
The proposed framework, depicted in Fig.~\ref{fig:Fig3}, is composed of three main modules: the \textit{knowledge-based reasoning engine}, the \textit{communication
layer}, and the \textit{smart motion planner}.

The \textit{knowledge-based reasoning engine} has two main components, the \textit{Abstract Knowledge} which represents the whole knowledge using an ontology
model\footnote{OWL files are accessible at: https://sir.upc.edu/projects/ontologies/.},
and the \textit{Reasoning Module} which infers ontological knowledge using Prolog predicates integrated within Knowrob.

The \textit{smart motion planning} module consists of the \textit{Instantiated Knowledge} module and the \textit{Planner} module.
The \textit{Instantiated Knowledge} module contains the dynamic knowledge that describes the temporary manipulation constraints (inferred form the abstract knowledge
based on the reasoning process), that are updated at each instance of time and that describes the way the bodies can be manipulated at a given particular instance of
time.
The \textit{Planner} module is composed of the \textit{The Kautham Project} \cite{Rosell2014}, an open-source tool for motion planing, developed in C++, that uses
the \textit{Open Motion Planning Library (OMPL)} \cite{sucanMK2012} as the core set of planning algorithms. OMPL allows planning under geometric constraints as well
as under differential constraints, including those that required dynamic
simulations (OMPL uses the \textit{Open Dynamic Engine}   for the dynamic simulation).

The role of the \textit{communication} layer is to transfer information between the other two layers. It includes ROS \cite{ros2009} and Json-prolog (which is
provided by Knowrob). The Json-prolog library enables the access to Prolog predicates through the ROS communication protocols, thus allowing to encapsulate the
knowledge-based reasoning process as a ROS service node that can be called from a ROS client node containing the motion planner. In this way, it is possible to
integrate the
Prolog-based reasoning process within the motion planner.

\subsection{Planning algorithm} \label{ss-Algorithm}

\begin{algorithm}
\begin{algorithmic}[1]
{\small
\REQUIRE Initial state $\mibf{q}_{\textrm{init}}$, Goal region $\mibf{Q}_{\textrm{goal}}\in {\cal C}$, Threshold $T_{max}$\\
\ENSURE A path from $\mibf{q}_{\textrm{init}}$ to $ \mibf{q} \in \mibf{Q}_{\textrm{goal}}$
\STATE ReadTheWorld()
\STATE K $\leftarrow$ FormulateOntology()
\WHILE {$t<T_{max}$}
\STATE  $\kappa \leftarrow$ KnowledgeReasoner($K,\kappa$)
\STATE SelectNodeToExpand()
\STATE  $u \leftarrow$ SampleControls()
\STATE  $\mibf{q}_{\textrm{new}}\leftarrow$ Propagate($u$,$\kappa$)
\IF {Valid( $\mibf{q}_{\textrm{new}}$) }
 \STATE UpdateConnections()
\IF {$\mibf{q}_{\textrm{new}} \in \mibf{Q}_{\textrm{goal}}$}
           \RETURN Path($\mibf{q}_{\textrm{new}}$)
\ENDIF
\ENDIF
\ENDWHILE
\RETURN {\small \sf NULL}
}
\end{algorithmic}
\caption{Ontological Physics-Based Motion Planning.}
\label{alg:OPBPlanner}
\end{algorithm}

The ontological physics-based planning procedure is sketched in Algorithm 1. It takes as input the initial configuration $\mibf{q}_{init}$, the goal region $\mibf{Q}_{goal}$ and a
time limit $T_{max}$, and returns a path from $\mibf{q}_{init}$ to $\mibf{q}_{goal}\in \mibf{Q}_{goal}$, if found, or {\small \sf NULL} otherwise.
The planning is performed by any of the OMPL control planners, like RRT, KPIECE, or PDST \cite{ladd2005}, tuned to use ODE as
state propagator. The instantiated knowledge is updated at each iteration and conditions the behavior of the bodies in the ODE world.

The steps of the algorithm are the following:
\begin{itemize}
\item \textit{ReadTheWorld}: Sets the initial state of the robot and the environment by reading the $p$, $o$, $v$, $w$, and $\eta$ of each body.
\item \textit{FormulateOntology}: Extracts the abstract knowledge $K$ about the world, such as the type of the bodies, the manipulation constraints and the
geometrical positions of bodies and determines whether the GoalRegion is occupied or not. Further it computes the ManipulationRegion for each constraint-oriented
manipulatable object. All these attributes are stored in the form of abstract knowledge.
\item \textit{KnowledgeReasoner}: Evaluates the state of the world and updates the instantiated knowledge $\kappa$. For instance,
if the GoalRegion is occupied by an object, then the sampling of the planner will be biased
towards one of the ManipulationRegion of the object occupying it (instead of the standard bias towards the goal). When the robot enters in the ManipulationRegion, the standard bias is reset and
the instantiated knowledge is updated with the manipulation constraints of that object.
\item \textit{SelectNodeToExpand}: Selects the node to expand  the tree data structure of the planner.
\item \textit{SampleControls}: Samples the controls within the given range.
\item \textit{Propagate}: Applies the controls to the bodies (controls can be applied by applying a force, a torque or by setting a linear and angular velocity),
while applying the controls, the state propagator will take into account the manipulation constraints provided by the instantiated knowledge, and returns the new
generated state.
\item \textit{Valid}: Returns true if no forbidden collision has occurred during the generation of the new state, and false otherwise.
\item \textit{UpdateConnections}: Updates the tree data structure by adding the edge between the previous and the newly generated state.
\item \textit{Path($\mibf{q}$)}: Returns a path from $\mibf{q}_{\textrm{init}}$ to $\mibf{q}\in\mibf{Q}_{\textrm{goal}}$ by backtracking along the planner data
strcuture.
\end{itemize}

\subsection{Simulation Results} \label{ss-Results}

\begin{figure}[t]
\begin{center}
    \includegraphics[width=0.91\columnwidth]{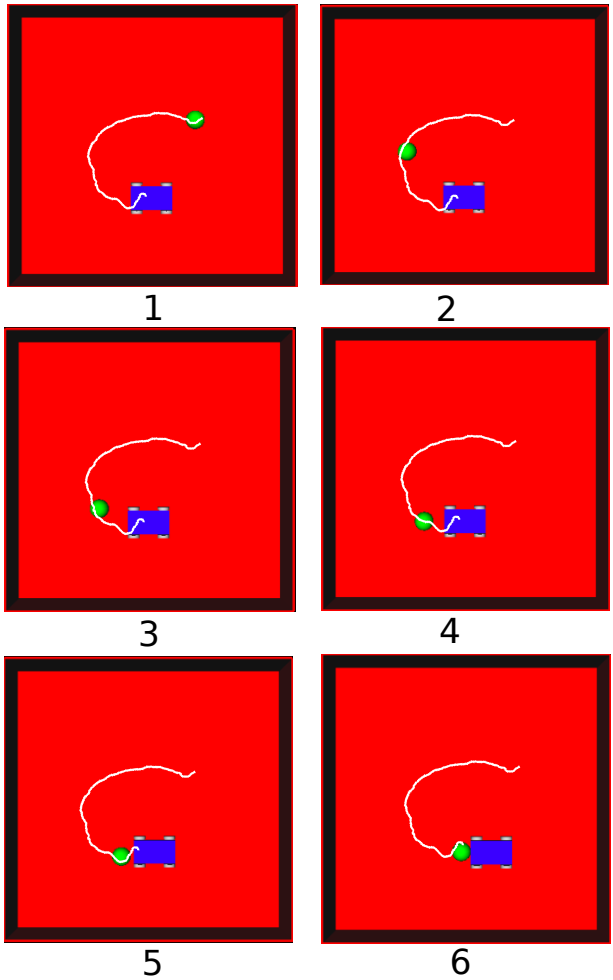}
\caption{Simulation results of the ontological physics-based motion planner using KPIECE.}\label{sim:KPIECE}
    \end{center}
\end{figure}

\begin{figure}[t]
\begin{center}
   \includegraphics[width=0.95\columnwidth]{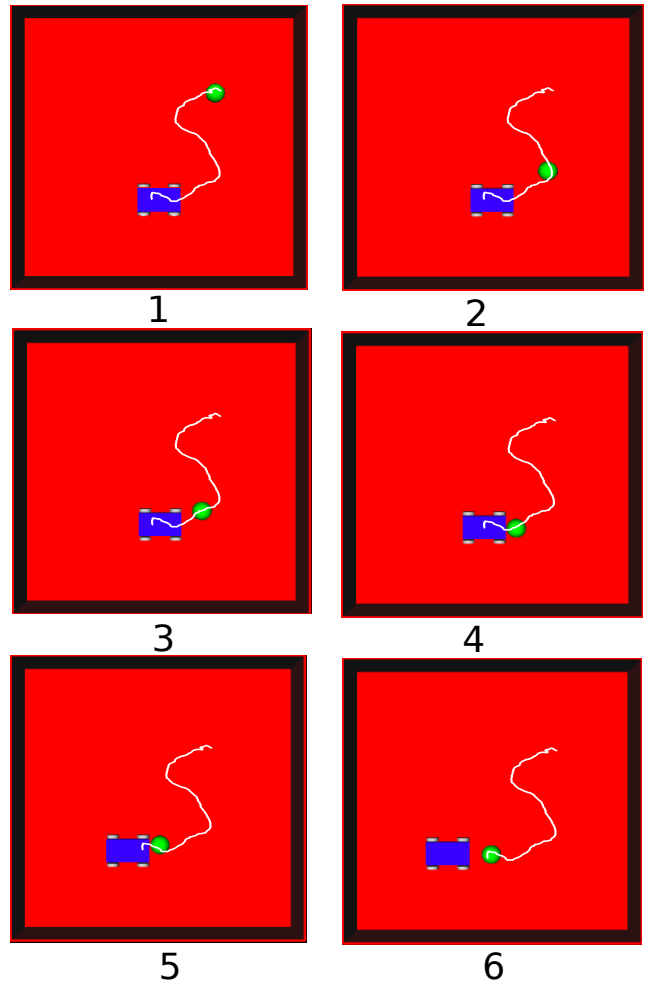}

\caption{Simulation results of the ontological physics-based motion planner using RRT.}\label{sim:RRT}
    \end{center}
\end{figure}

\begin{figure}[t]
\begin{center}
   \includegraphics[scale=.42]{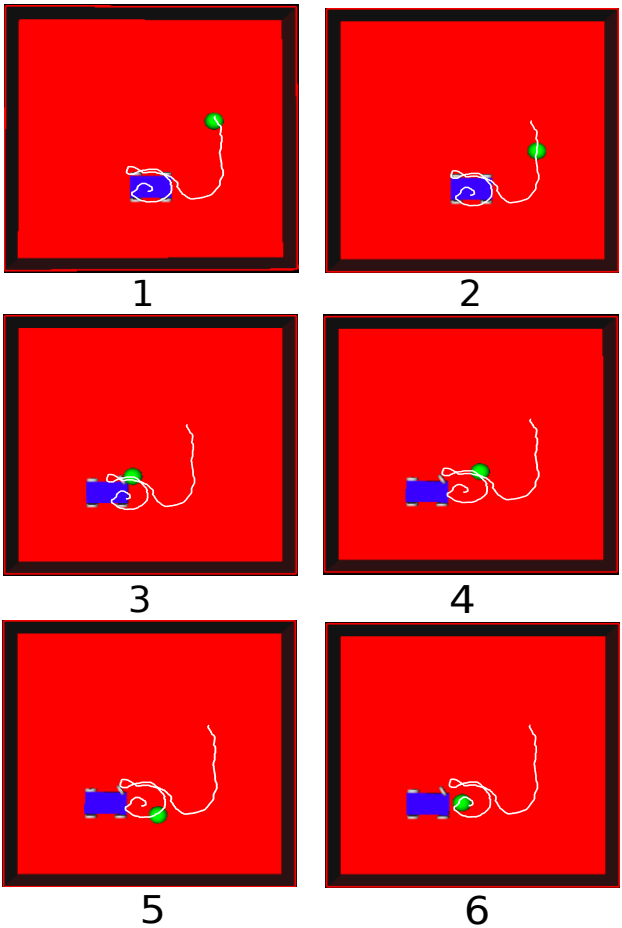}

\caption{Simulation results using KPIECE without using instantiated knowledge.}\label{sim:Simple-Phy-Based}
\end{center}
\end{figure}

\begin{figure}[t]
\begin{center}
\centering
   \includegraphics[width=8cm]{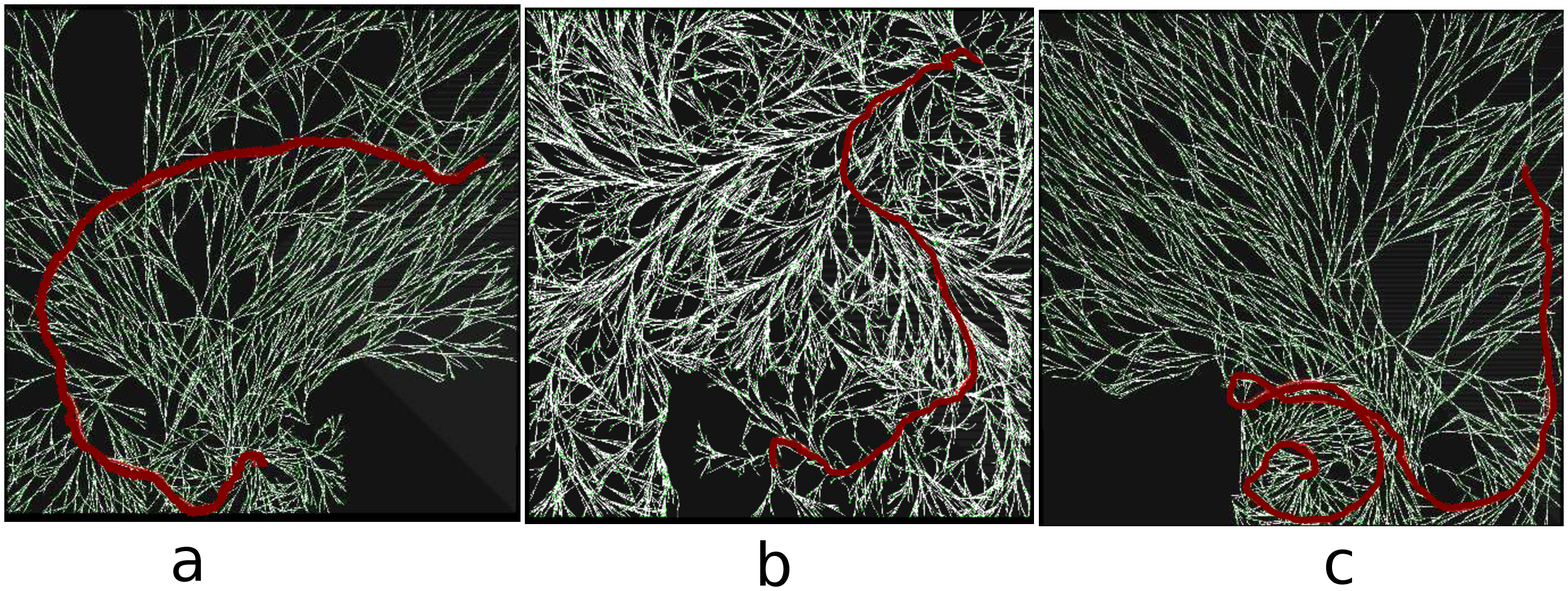}

\caption{Configuration spaces with the tree structures of the planners used and the corresponding solution paths (in red): a) KPIECE using instantiated knowledge; b) RRT using instantiated knowledge; c) KPIECE without using instantiated knowledge.}\label{sim:CSpace}
\end{center}
\end{figure}

 A simple scenario has been set in order to put the focus on how the planner makes use of the knowledge to manage a complex situation involving a manipulatable obstacle. It consists of: a) the walls of a room which are defined as fixed bodies; b) a spherical shaped robot with two degrees of freedom; and c) a car-like obstacle which
 is defined as a constraint-oriented manipulatable body that can only be moved in the forward and backward
 directions. The goal region $\mibf{Q}_{goal}$  is occupied by the car-like obstacle, so no collision-free path exists, i.e. the robot needs to push the car-like
 obstacle away in order to reach $\mibf{Q}_{goal}$.

 The knowledge-based reasoning engine reads the initial state of the world and extracts the abstract knowledge related to the world such as the type of bodies, the
manipulation constraints, and the state of the goal. Based on a reasoning process, the instantiated knowledge is inferred from the abstract
knowledge and fills the data structures in the motion planning layer (such as the current manipulation constraints, the valid manipulation regions, the goal state - occupied
or free -, etc.), that are periodically updated. In this example, the knowledge-based reasoning engine determines that the goal is occupied and extracts the manipulation constraints
associated to the car-like obstacle (they have been defined as two ManipulationRegions corresponding to its front and
rear parts, since contact is allowed with the front and the rear sides and is forbidden with the two lateral sides).

The planning of paths has been performed  using two different physics-based kinodynamic motion planners, KPIECE and RRT.
At each instance of time, the instantiated knowledge evaluates the state of the world and the manipulation constraints, changing the standard sampling bias towards the goal by a bias towards one of the ManipulationRegions of the car-like obstacle that occupies the GoalRegion. Once the robot reaches the ManipulationRegion, the standard bias towards the goal is restored.
Fig.~\ref{sim:KPIECE} and~\ref{sim:RRT} show, respectively, sequences of snapshots of the solutions found using  KPIECE and RRT.
In both cases, it can be appreciated how the robot moves towards one of the ManipulationRegions and then hits the car-like obstacle at its front/rear side and pushes it to free the GoalRegion and reach the goal.

 Without using the aid provided by the instantiated knowledge, these planners were not able to find good solutions, as illustrated in Fig.~\ref{sim:Simple-Phy-Based}.
 Since in this case the planner did not know how to manipulate the car-like object, many tree edges tried unfruitfully to grow towards the goal by hitting the car at its side
 (in the solution shown the robot hits the car at the wheel and bounces back). Therefore, the computational cost was greater (it took approximately twice the time used by the ontological physics-based planners, either with KPIECE or RRT).
The configuration spaces for all the above stated executions are shown in Fig. \ref{sim:CSpace}.

\section{Conclusions and future work}\label{s-Conclusions}

The present study has explored the integration of knowledge-based reasoning with physics-based motion planning.
A framework has been presented to combine both phases, aiming to enhance the planning process for manipulation problems. Two types of reasoning
(about manipulation actions and about the geometry of objects) is considered. The result of the inference process is
stored inside the instantiated knowledge and provided to the motion planning. The proposed algorithm has been illustrated
 with a manipulation problem where the goal region is occupied by a manipulatable obstacle that should be removed by the robot using
  pushing actions. The results described show that the proposed algorithm has a better performance than the simple physics-based planning.


Future work will be directed towards the integration of physics-based reasoning with a task planner, and the consideration of other manipulation actions besides pushing, like pick and place actions.

%
%

\balance
\bibliographystyle{IEEEtran}
\bibliography{References}

\begin{thebibliography}{10}
\providecommand{\url}[1]{#1}
\csname url@samestyle\endcsname
\providecommand{\newblock}{\relax}
\providecommand{\bibinfo}[2]{#2}
\providecommand{\BIBentrySTDinterwordspacing}{\spaceskip=0pt\relax}
\providecommand{\BIBentryALTinterwordstretchfactor}{4}
\providecommand{\BIBentryALTinterwordspacing}{\spaceskip=\fontdimen2\font plus
\BIBentryALTinterwordstretchfactor\fontdimen3\font minus
  \fontdimen4\font\relax}
\providecommand{\BIBforeignlanguage}[2]{{%
\expandafter\ifx\csname l@#1\endcsname\relax
\typeout{** WARNING: IEEEtran.bst: No hyphenation pattern has been}%
\typeout{** loaded for the language `#1'. Using the pattern for}%
\typeout{** the default language instead.}%
\else
\language=\csname l@#1\endcsname
\fi
#2}}
\providecommand{\BIBdecl}{\relax}
\BIBdecl

\bibitem{Perez1983}
T.~Lozano-P\'erez, ``{Spatial Planning: A Configuration Space Approach.}''
  \emph{IEEE Trans. on Computers}, vol.~32, no.~2, pp. 108--120, 1983.

\bibitem{sucanK2012}
I.~A. {\c{S}}ucan and L.~E. Kavraki, ``A sampling-based tree planner for
  systems with complex dynamics,'' \emph{IEEE Trans. on Robotics}, vol.~28,
  no.~1, pp. 116--131, 2012.

\bibitem{lavalle2000}
S.~M. Lavalle and J.~J. Kuffner, ``{Rapidly-Exploring Random Trees: Progress
  and Prospects},'' B.~R. Donald, K.~M. Lynch, and D.~Rus, Eds.\hskip 1em plus
  0.5em minus 0.4em\relax A K Peters, 2001, pp. 293--308.

\bibitem{csucan2010}
I.~A. {\c{S}}ucan and L.~E. Kavraki, ``Kinodynamic motion planning by
  interior-exterior cell exploration,'' in \emph{Algorithmic Foundation of
  Robotics VIII}.\hskip 1em plus 0.5em minus 0.4em\relax Springer, 2010, pp.
  449--464.

\bibitem{donald1993}
B.~Donald, P.~Xavier, J.~Canny, and J.~Reif, ``Kinodynamic motion planning,''
  \emph{Journal of the ACM}, vol.~40, no.~5, pp. 1048--1066, 1993.

\bibitem{zickler2010}
S.~Zickler and M.~M. Veloso, ``Variable level-of-detail motion planning in
  environments with poorly predictable bodies.'' in \emph{Proc. of the European
  Conf. on Artificial Intelligence Montpellier}, 2010, pp. 189--194.

\bibitem{OpenDE2007}
S.~Russell, ``Open {D}ynamic {E}ngine,'' http://www.ode.org/, 2007.

\bibitem{zickler2009}
S.~Zickler and M.~Veloso, ``Efficient physics-based planning: sampling search
  via non-deterministic tactics and skills,'' in \emph{Proc. of The 8th Int.
  Conf. on Autonomous Agents and Multiagent Systems-Volume 1}, 2009, pp.
  27--33.

\bibitem{mosenlechner2013}
L.~Mosenlechner and M.~Beetz, ``Fast temporal projection using accurate
  physics-based geometric reasoning,'' in \emph{Proc. of the IEEE Int. Conf. on
  Robotics and Automation}, 2013, pp. 1821--1827.

\bibitem{dogar2012}
M.~R. Dogar, K.~Hsiao, M.~Ciocarlie, and S.~Srinivasa, ``Physics-based grasp
  planning through clutter,'' in \emph{Robotics: Science and Systems}, 2012.

\bibitem{kitaev2015}
N.~Kitaev, I.~Mordatch, S.~Patil, and P.~Abbeel, ``Physics-based trajectory
  optimization for grasping in cluttered environments,'' in \emph{Proc, of the
  IEEE Int. Conf. on Robotics and Automation}, 2015, pp. 3102 -- 3109.

\bibitem{feyzabadi2014}
S.~Feyzabadi and S.~Carpin, ``Knowledge and data representation for motion
  planning in dynamic environments,'' in \emph{Robot Intelligence Technology
  and Applications 2}.\hskip 1em plus 0.5em minus 0.4em\relax Springer, 2014,
  pp. 233--240.

\bibitem{TaskPlanning2015}
A.~Akbari, Muhayyudin, and J.~Rosell, ``Task and motion planning using
  physics-based reasoning,'' in \emph{Proc. of the IEEE Int. Conf. on Emerging
  Technologies and Factory Automation}, 2015.

\bibitem{owl2004}
G.~Antoniou and F.~van Harmelen, ``Web {O}ntology {L}anguage: {OWL},'' in
  \emph{Handbook on Ontologies in Information Systems}, S.~Staab and R.~Studer,
  Eds.\hskip 1em plus 0.5em minus 0.4em\relax Springer-Verlag, 2003, pp.
  67--92.

\bibitem{protege}
Stanford2007, ``Prot\'eg\'e,'' http://protege.stanford.edu/, 2007.

\bibitem{knowrob2009}
M.~Tenorth and M.~Beetz, ``Knowrob knowledge processing for autonomous personal
  robots,'' in \emph{Proc. of the IEEE/RSJ Int. Conf. on Intelligent Robots and
  Systems}, 2009, pp. 4261--4266.

\bibitem{Rosell2014}
J.~Rosell, A.~P\'erez, A.~Aliakbar, Muhayyuddin, L.~Palomo, and N.~Garc\'{\i}a,
  ``The kautham project: A teaching and research tool for robot motion
  planning,'' in \emph{Proc. of the IEEE Int. Conf. on Emerging Technologies
  and Factory Automation}, 2014.

\bibitem{sucanMK2012}
I.~A. {\c{S}}ucan, M.~Moll, and L.~E. Kavraki, ``The {O}pen {M}otion {P}lanning
  {L}ibrary,'' \emph{{IEEE} Robotics \& Automation Magazine}, no.~4, pp.
  72--82, December.

\bibitem{ros2009}
M.~Quigley, K.~Conley, B.~Gerkey, J.~Faust, T.~Foote, J.~Leibs, R.~Wheeler, and
  A.~Y. Ng, ``{ROS}: an open-source robot operating system,'' in \emph{ICRA
  {W}orkshop on {O}pen {S}ource {S}oftware}, vol.~3, no. 3.2, 2009, p.~5.

\bibitem{ladd2005}
A.~Ladd and L.~Kavraki, ``Fast tree-based exploration of state space for robots
  with dynamics,'' in \emph{Algorithmic Foundations of Robotics VI}.\hskip 1em
  plus 0.5em minus 0.4em\relax Springer, 2005, pp. 297--312.

\end{thebibliography}
\nocite{}
\end{document}